\documentclass[runningheads]{llncs}
\usepackage{graphicx}

\usepackage{siunitx}
\usepackage{booktabs}
\usepackage{tabularx}
\usepackage{makecell}
\usepackage{colortbl}

\newcolumntype{C}{>{\centering\arraybackslash}X}
\newcolumntype{R}{>{\raggedleft\arraybackslash}X}
\newcolumntype{L}{>{\raggedright\arraybackslash}X}
\newcolumntype{P}[1]{>{\raggedleft\arraybackslash}p{#1}}
\usepackage{xcolor}
\usepackage{hyperref}
\usepackage{enumitem}
\usepackage{tablefootnote}
\usepackage{multirow}
\usepackage{listings}
\usepackage{float}
\usepackage{subcaption}
\usepackage{cite}
\usepackage{listings}
\usepackage{orcidlink}

\newcommand{\tabularxmulticolumncentered}[3]{\multicolumn{#1}{>{\centering\hsize=\dimexpr#1\hsize+#1\tabcolsep+\arrayrulewidth\relax}#2}{#3}}
\newcommand{\colheader}[1]{\tabularxmulticolumncentered{3}{C|}{#1}}
\newcommand{\rowheader}[2]{\parbox[t]{8mm}{\multirow{#1}{*}{\rotatebox[origin=c]{90}{\shortstack{#2}}}}}

\newcounter{nalg}[chapter] 
\renewcommand{\thenalg}{\thechapter .\arabic{nalg}} 
\DeclareCaptionLabelFormat{algocaption}{Algorithm \thenalg} 

\lstnewenvironment{algorithm}[1][] 
{   
    \refstepcounter{nalg} 
    \captionsetup{labelformat=algocaption,labelsep=colon} 
    \lstset{ 
        mathescape=true,
        frame=tB,
        numbers=left, 
        numberstyle=\tiny,
        basicstyle=\scriptsize, 
        keywordstyle=\color{black}\bfseries\em,
        keywords={,input, output, return, datatype, function, in, if, else, foreach, while, begin, end, } 
        numbers=left,
        xleftmargin=.04\textwidth,
        #1 
    }
}
{}

\usepackage{amssymb}

\makeatletter
\newcommand{\hlinex}[1]{%
    \noalign {\ifnum 0=`}\fi \hrule height #1
    \futurelet \reserved@a \@xhline
}
\newcolumntype{"}{@{\hskip\tabcolsep\vrule width 1pt\hskip\tabcolsep}}
\makeatother

\newfloat{lstfloat}{tb}{lop}
\floatname{lstfloat}{Listing}

\newcommand{\ablationresultmarker}[1]{%
  \begingroup\normalfont
  \includegraphics[height=1.4\fontcharht\font`\W]{figures/ablation/markers/#1}%
  \endgroup
}

\begin{document}
\title{A Universal Prompting Strategy for Extracting Process Model Information from Natural Language Text using Large Language Models}

\titlerunning{Extracting Process Model Information Using LLMs}

\author{%
    Julian Neuberger\inst{1}\orcidlink{0009-0008-4244-7659} 
    \and Lars Ackermann\inst{1}\orcidlink{0000-0002-6785-8998} 
    \and Han van der Aa\inst{2}\orcidlink{0000-0002-4200-4937}
    \and Stefan Jablonski\inst{1}
}

\authorrunning{J. Neuberger et al.}
%
\institute{%
    University of Bayreuth, Bayreuth, Germany\\
    \email{firstname.lastname@uni-bayreuth.de}
\and
    University of Vienna, Vienna, Austria\\
    \email{han.van.der.aa@univie.ac.at}
}
\maketitle
\begin{abstract}

Over the past decade, extensive research efforts have been dedicated to 
the extraction of information from textual process descriptions. Despite 
the remarkable progress witnessed in natural language processing (NLP), information extraction within the Business 
Process Management domain remains predominantly reliant on rule-based 
systems and machine learning methodologies. 
Data scarcity has so far prevented the successful application of deep learning techniques.
However, the rapid progress in generative large language models (LLMs) makes it possible to solve many NLP tasks with very high quality without the need for extensive data. 
Therefore, we systematically investigate the potential of LLMs for extracting information from textual process descriptions, targeting the detection of process elements such as activities and actors, and relations between them. Using a heuristic algorithm, we demonstrate the suitability of the extracted information for process model generation.
Based on a novel prompting strategy, we show that LLMs are able to outperform
state-of-the-art machine learning approaches with absolute performance 
improvements of up to 8\% $F_1$ score across three different datasets.
We evaluate our prompting strategy on eight different LLMs, showing
it is universally applicable, while also analyzing the impact of certain prompt
parts on extraction quality. The number of example texts, the specificity of 
definitions, and the rigour of format instructions are identified as key for 
improving the accuracy of extracted information. Our code, prompts, and data 
are publicly available\footnote{
See \url{https://github.com/JulianNeuberger/llm-process-generation/tree/er2024}
and \url{https://github.com/JulianNeuberger/pet-to-bpmn-poc}}.


\end{abstract}
\section{Introduction}\label{sec:intro}
In the field of Business Process Management (BPM), process models are established tools for designing, implementing, enacting, and analyzing enterprise processes~\cite{davies2006practitioners}. However, the manual creation of these models is very time-consuming and accounts for around 60\% of the total time spent on process management~\cite{friedrich2011process}. In order to reduce this effort, the automatic creation of these models based on a variety of information sources is a research focus in the field of BPM~\cite{van2016processmining,friedrich2011process}. In this respect, the paper at hand contributes to the extraction of process-relevant information from natural language information sources.

Information on organizational processes is frequently contained in a range of textual documents, such as process descriptions, rules and regulations, and work instructions~\cite{van2018challenges,van2017causes}. 
Recognizing this, a variety of techniques has been developed that aim to automatically extract process information from texts in order to subsequently turn it into process models~\cite{friedrich2011process,bellan2022extracting,van2019extracting}. This two-step procedure, in which information is extracted first and turned into a process model second, comes with several advantages in comparison to a direct text-to-model transformation approach: (1) The result quality can be evaluated with established means from the information-extraction domain, (2) extracted information can be transformed in more than one target process modeling language\footnote{This is inspired by the paradigm of \textit{interlingua-based machine translation}~\cite{richens1958interlingual}, which reduces the number of translation systems for n languages from $n^2$ to $2n$.}, and (3) it is possible to use extracted process information for other purposes such as, for instance,  compliance checking, formal reasoning~\cite{van2018checking,quishpi2020extracting}, and process querying~\cite{leopold2019searching}. 

\begin{figure}[tb]
    \centering
    \includegraphics[width=\textwidth]{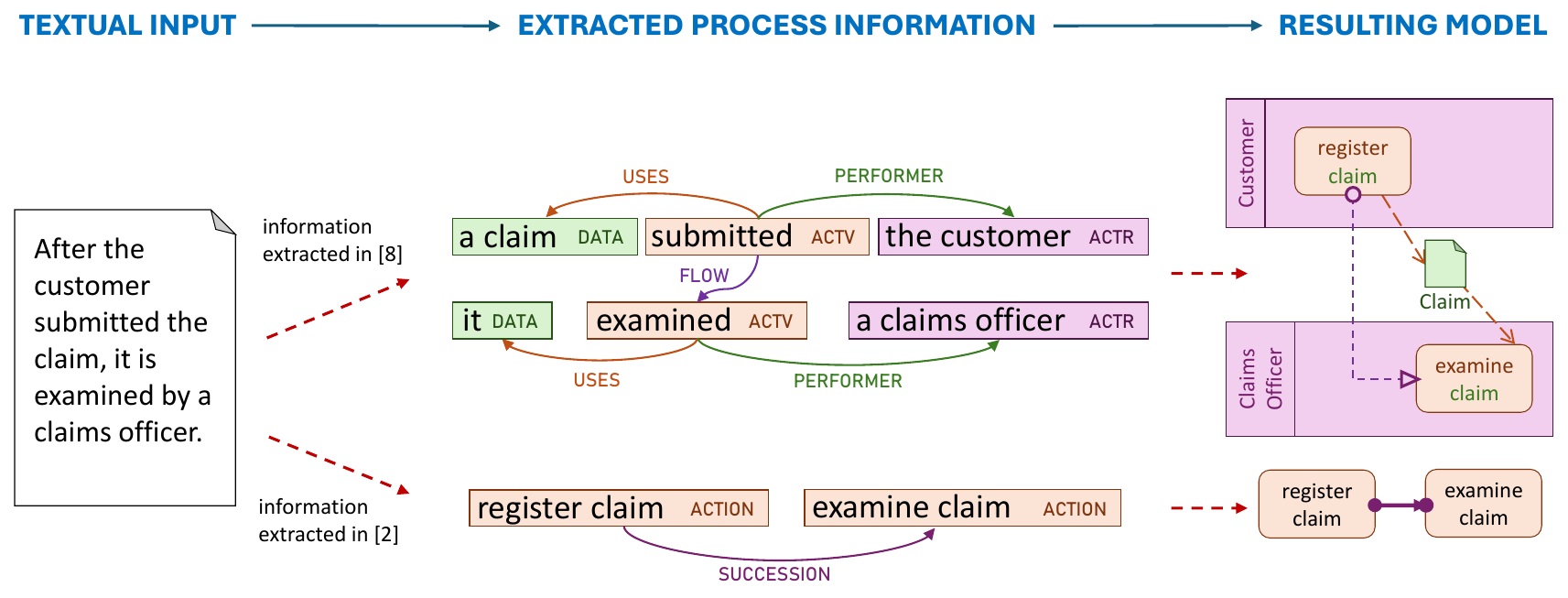}
    \caption{Fragment of a larger text describing a business process of an 
    insurance company. Different methodologies may extract different process 
    relevant information, depending on the target modelling notation or use case.
    }
    \label{fig:running-example}
\end{figure}

The goal, scope, and challenges of information extraction depend on the input document type and content, as well as the desired output, i.e., the information to be extracted.
Still, the extraction of process information from text generally involves:
(1) the identification of textual mentions of process entities, such as activities, process participants, and business objects, and (2) relations between these entities, such as sequential dependencies, exclusivity, and assignments (e.g., who performs which step). 
As an illustration, 
Figure~\ref{fig:running-example} shows a fragment of a textual description and two instantiations of the extraction task, focused on the information necessary for a model in Business Process Model and Notation (BPMN)\footnote{\url{https://www.omg.org/bpmn/}, accessed June 2, 2024, also see Section~\ref{sec:model-generation}.} model~\cite{bellan2023pet} (upper part) and for declarative process modeling~\cite{van2019extracting} (lower part). As shown, they involve different entities and relations, which each need to be inferred from the unstructured textual input.

A key problem is that the extraction of process information is still largely rule-based~\cite{neuberger2023beyond}. However, crafting useful rules is complicated, requires an 
extensive understanding of the process itself, and the rules are hard to 
transfer across organizations or text sources. To overcome this, recent work proposed the use of machine learning techniques~\cite{neuberger2023beyond}, though these are hampered by data scarcity. Work that strives towards using pre-trained generative LLMs, e.g., GPT-3~\cite{bellan2022extracting} aims to circumvent this concern.
However, the work in~\cite{bellan2022extracting} only presents a preliminary study,
with limitations in terms of analyzed datasets, extracted information, and 
result discussion. 
Therefore, this paper aims to provide deeper insights into the usability of LLMs for process information extraction and specifically includes the following core contributions:  
\textbf{(I)} It presents challenges that make the extraction of process-relevant information in particular a difficult task (Section~\ref{sec:challenges}). 
\textbf{(II)} As our main contribution, it proposes a novel, task-specific, and rigorously empirically validated prompting strategy for solving the aforementioned information extraction tasks (Section~\ref{sec:prompting}). 
\textbf{(III)} For transforming the extracted information into a process model, we propose a proof-of-concept algorithm, illustrating that the information extracted is useful for automated process model generation (Section~\ref{sec:model-generation}). 
\textbf{(IV)} It provides the currently most comprehensive study of using LLMs for extracting business process relevant information from natural language text (Sections~\ref{sec:experiment_setup} and ~\ref{sec:results}). To this end we rigorously compare our prompting strategy on multiple datasets with state-of-the-art approaches and achieve up to 7\% higher absolute $F_1$ scores compared to machine learning methods and up to 8\% compared to rule-based methods. 
\textbf{(V)} By testing our prompting strategy with eight state-of-the-art LLMs, we empirically demonstrate the generality of both our results and the applicability of our prompting strategy.
\textbf{(VI)} An ablation study (Section~\ref{sec:ablation}) shows that common best practices in prompt engineering are only of limited use for process information extraction. Thus, we also define guidelines for using LLMs for process information extraction (Section~\ref{sec:lessons-learned}).

\noindent The rest of this paper is structured as follows. 
Section~\ref{sec:challenges} describes the information extraction tasks and its challenges in detail. Section~\ref{sec:related_work} summarizes the current state of the art in dealing with these tasks. After that we, describe our prompting strategy, a model generation algorithm, the experiments, and corresponding results (Section~\ref{sec:approach}--\ref{sec:results}). Section~\ref{sec:conclusion} describes limitations and future work.  

\section{Task Descriptions and Challenges}\label{sec:challenges}

In this section, we describe the three main (sub)tasks of process information extraction (Section~\ref{sec:task-descriptions}), before highlighting a range of challenges associated with such extraction  and with the use of LLMs for it (Sections \ref{sec:challenges-task}--\ref{sec:challenges-llm}).


\subsection{Task descriptions}\label{sec:task-descriptions}

Our work focuses on three established subtasks of (process) information extraction from text: Mention Detection (MD), Entity Resolution (ER), and Relation Extraction (RE)~\cite{ackermann2023bridging,van2019extracting,quishpi2020extracting,neuberger2023beyond}.


\noindent \textbf{Mention Detection}\label{task:md} (MD) is concerned with finding and extracting text fragments that contain process relevant information, such as activities (or actions), process-relevant objects or data (i.e., business objects), or involved persons and departments (i.e., actors). For instance, in Figure~\ref{fig:running-example}, the upper example shows mentions of \textit{data}, \textit{actions}, and \textit{actor}, whereas the lower one focuses on \textit{activities}.
This definition is similar to Named Entity Recognition (NER), 
though we also extract spans not covered by the traditional 
definition of NER, e.g., activities.


\noindent \textbf{Entity Resolution}\label{task:er} (ER) aims to recognize when different mentions refer to the same process entity.
For example, in Figure~\ref{fig:running-example}, successful \hyperref[task:er]{ER} would identify that the word \textit{it} in ``it is examined'' corresponds to the \textit{claim} mentioned in the previous phrase. Another common example is using  \hyperref[task:er]{ER} to recognize that the same actor (across mentions) performs different steps.
\hyperref[task:er]{ER} is a super-set of co-reference
resolution and anaphora resolution~\cite{sukthanker2020anaphora} and is a crucial step when dealing with process-related texts, which frequently involve repeated mentions across sentences or even paragraphs~\cite{neuberger2023beyond}. 

\noindent \textbf{Relation Extraction}\label{task:re}\label{task:ce} (RE) is the task of
detecting and classifying relations between mentions. 
Relations are usually directed and have one (unary relation), or two (binary relations) arguments.
For instance, the upper example in Figure~\ref{fig:running-example} shows three kinds of relations: \textit{uses} signals which data objects are used by an activity, \textit{performer} captures which actor performed an activity, and \textit{flow} captures a sequential relation between two activities.
\hyperref[task:re]{RE} is crucial when it comes to information extraction in our context, given that processes inherently involve process steps (i.e., activities) that are connected to each other through relations.
Note that we regard constraint extraction~\cite{van2019extracting,quishpi2020extracting} (CE), which relates to declarative process modeling, as an \hyperref[task:re]{RE} problem: constraints have one or two arguments,
are directed, and carry type information (e.g., \textit{Succession}, \textit{Init}).


\subsection{Challenges of Process Information Extraction from Text}
\label{sec:challenges-task}
Information extraction, a common task in natural language processing (NLP), faces general challenges, which are also well-known in BPM literature~\cite{van2018challenges,franceschetti2023ambiguity}, and often 
central elements of interest in the design of rule-based and learning-based systems alike~\cite{van2019extracting,neuberger2023beyond}. Simply using LLMs for process information extraction solves some of these challenges and justifies the investigation of their applicability.

In the context of process-related texts \emph{\textbf{Linguistic Variance}} means that the same behavior or process characteristics can be described in a variety of ways such as, for instance, active and passive voice.
%
\emph{\textbf{Context Cues}} are a challenge in that single words can fundamentally alter the meaning of a process description (e.g., \emph{``first, a claim is created'' and inverted semantics in ``finally, a claim is created''}.
%
Processes are typically described in sequential form, although they usually contain branches (e.g. XOR decision branches). This results in \emph{\textbf{Long-distance Relations}} that existing approaches struggle with~\cite{neuberger2023beyond} or cannot handle~\cite{van2019extracting}.
\emph{\textbf{Implicit and Ambiguous Information}} such as the ``examination target'' in \emph{``after registering the file in the database, it needs to be examined''} needs to be interpreted~\cite{van2018checking,franceschetti2023ambiguity}.
Research indicates a negative correlation between correctness of extracted process information and \emph{\textbf{Text Length}}~\cite{ackermann2023bridging}.
Finally, the application of deep learning is hindered by the fact that the largest available data set contains only 45 human-annotated process descriptions~\cite{bellan2023pet} (\emph{\textbf{Small Datasets}}).

LLMs are able to overcome the above challenges~\cite{wei2022chain}, which is why this paper analyzes their suitability for process information extraction, as proposed in previous work~\cite{bellan2022extracting,ackermann2021data}. 
However, LLMs require great care in the formulation of the input (prompts)~\cite{bellan2022extracting,wei2022chain,white2023promptpatterncatalog,tornberg2024bestpractices,min2023recent}. In \cite{min2023recent} authors argue: ``a good prompt can be worth hundreds of labeled data points''. For this reason, the core of the present work lies in the development (Section~\ref{sec:prompting}) and evaluation (Sections~\ref{sec:experiment_setup} and \ref{sec:ablation}) of suitable prompts for process information extraction.

\subsection{Challenges of Process Information Extraction Using LLMs}\label{sec:challenges-llm}

Using LLMs for process information extraction from texts helps with linguistic challenges, but adds itself several additional challenges. We discuss these 
here and reference them later in Section~\ref{sec:lessons-learned} to show
how we can deal with them.

\textbf{(\textit{C1}) Limited output control}\label{challenge:output-control}. Input and output consist of plain text. Given that the input for inference is raw text, it inherently suits LLMs for our tasks (cf. Section~\ref{sec:task-descriptions}). However, as the expected output should adhere to a specific schema, it becomes necessary to instruct the LLM to conform to this schema. Moreover, this principle necessitates a robust output parser, as LLMs tend to exhibit variability in their output, which presently cannot be entirely eradicated. Having only limited control
over generated output is especially problematic for the BPM domain, where definitions for
relevant information often overlap, e.g., \textit{actions} (just predicate) versus
\textit{activities} (predicate and object).

\textbf{(\textit{C2}) Input presentation dependencies}\label{challenge:input-presentation}. Although LLMs provide an interface for natural language input, the quantity, form and level of detail must be carefully matched to the task at hand. The LLM faces the challenge of determining the importance of the input components. Furthermore, while LLMs emulate human reasoning, the interpretation of inputs may diverge significantly from that of human beings, thereby rendering prompt optimization a trial-and-error process. This challenge is further aggravated by some elements of process models, that
are complex to explain concisely, e.g., parallel and exclusive workflows.

\textbf{(\textit{C3}) Black-box}\label{challenge:black-box}. Deep learning methods generally suffer from challenges concerning
explainability of predictions~\cite{xu2019explainable}, which is also true for LLMs. Contrary
to classical learning methods, such as decision trees, LLMs offer no fail safe 
mechanism to validate extraction rules. This is problematic for business
process information extraction in particular, since recent work focuses on 
``human-in-the-loop'' systems~\cite{ter2023process}, where the human must be able to 
follow system decisions.

\textbf{(\textit{C4}) Data unawareness}\label{challenge:data-unawareness}. In contrast to generative AI models trained on task-specific data, an LLM is usually not aware of the particular dataset it is tasked to process. Thus, using LLMs to process a particular dataset requires to form instructions that precisely describe all relevant details of a dataset. The generalizing capabilities of LLMs can be
an additional hurdle in this context, especially, when declarative process models are 
concerned, where a multitude of constraint types exist. The LLM is likely to know of these 
through the pre-training process, and therefore may extract irrelevant ones for a given dataset.

\textbf{(\textit{C5}) Costly experiments}\label{challenge:costly-experiments}. Applying LLMs usually requires usage of commercial APIs (e.g., OpenAI), which come with downsides: (i) a token limit restricting the maximum input and output size and (ii) fees based on the number of tokens processed. In view of the many possible variations in the influencing parameters, conducting experiments can be cost-intensive. This
is especially true for the BPM domain, where the density of information in process descriptions
is very high, needing many output tokens to extract and encode it.
\section{Related Work}\label{sec:related_work}
Related approaches are divided into rule-based, machine learning (ML)-based, deep learning-based, and LLM-based process information extraction. An approach is considered related if it solves at least one of the tasks in Section~\ref{sec:task-descriptions}. 

\textbf{Rule-based approaches.}
Rule-based approaches leverage linguistic features to extract information from natural language process descriptions through explicitly coded mapping rules. For instance, Friedrich et al.'s seminal work~\cite{friedrich2011process} employs syntax features and word information from a lexical database to identify patterns at both sentence and document levels for BPMN model creation. Other approaches like those in \cite{sanchez2021unleashing,quishpi2020extracting} adopt similar techniques for automatic text annotation, employing regular expressions for syntactic dependency trees, and part-of-speech tags, showcasing superior performance on novel datasets. Additionally, \cite{van2019extracting} presents a rule-based technique, currently leading in extracting declarative process models from raw text using syntax parsing and word-level features. Similar advancements are seen in \cite{ferreira2017semi} and \cite{bellan2023pet}, the latter integrating ML-based entity \hyperref[task:md]{MD} with subsequent rule-based \hyperref[task:re]{RE}. Furthermore, \cite{lopez2018process} focuses on extracting Dynamic Condition Response (DCR) graphs. Recent studies suggest that while rule-based approaches can be tailored to specific tasks and data sets, they can hardly deal with ambiguity and linguistic variance.~\cite{bellan2022extracting,neuberger2023beyond}. 

\textbf{ML-based approaches.}
\cite{neuberger2023beyond} presents a ML extraction pipeline based on \cite{bellan2023pet} and is used as a baseline for our comparative evaluation (Section~\ref{sec:results}). 
The deep-learning approach presented in \cite{Qian2020} classifies text fragments analyzing the input text on several levels of granularity. However, extracting these fragments is not part of the approach, which simplifies the task of \hyperref[task:md]{MD} to mention classification, i.e., locating the information to extract is omitted. Though the work presented in \cite{ackermann2021data} overcomes this limitation and outperforms the approach, it does not support \hyperref[task:re]{RE}.  
In general, techniques of this paradigm either struggle to deal with linguistic variability and ambiguity, or they require vast amounts of training data, making them particularly unfeasible for small datasets (see Section~\ref{sec:challenges-task}).

\textbf{LLM-based approaches.}
Bellan et al.~\cite{bellan2022extracting} utilize pre-trained LLMs to cope with data scarcity, yet their approach exhibits three primary weaknesses: \textit{(i)} It is restricted to a subset of entity types, namely \textit{activities}, \textit{participants}, a \textit{performs} relation, and a \textit{direct-consequences} relation, \textit{(ii)} it lacks strict output formatting, hindering automated result processing
, unlike our prompting strategy
, and \textit{(iii)} its evaluation is limited to 7 of the 45 process descriptions from the \hyperref[data:pet]{PET} dataset, whereas we evaluate our modular prompt on the entire dataset, plus two validation datasets. Thus, direct comparison between \cite{bellan2022extracting} and our work is not feasible. Nonetheless, as in \cite{bellan2022extracting}, our modular prompt also descriptive instructions with input examples accompanied by their expected outputs. Although process models are generated using LLMs in \cite{kourani2024process}, the work is not comparable to ours, as \cite{kourani2024process} requires human involvement and only supports the extraction of activities and their arrangement in a directed graph (e.g., actors and data are missing).

\section{LLM-based Process Model Extraction from Text}\label{sec:approach}

To generate process models from natural language texts, we first extract process information using LLMs and a novel prompting strategy (Section~\ref{sec:prompting}). Next, we describe a baseline algorithm for generating a process model (Section~\ref{sec:model-generation}). 

\subsection{Process Information Extraction with LLMs}\label{sec:prompting}

Extracting process information with LLMs requires a \emph{prompt design} that addresses the challenges mentioned in Section~\ref{sec:challenges-llm}. Thus, a prompt structure consisting of the three modules \emph{Context}, \emph{Task Description} and \emph{Restrictions} is described below at the example of the Mention Detection task. However, the prompting strategies for the remaining tasks (Section~\ref{sec:task-descriptions}) are analogous. 

\subsubsection{High-Level Prompt Structure.}\label{sec:prompting-structure}
\begin{figure}[bt]
    \centering
    \includegraphics[width=\textwidth]{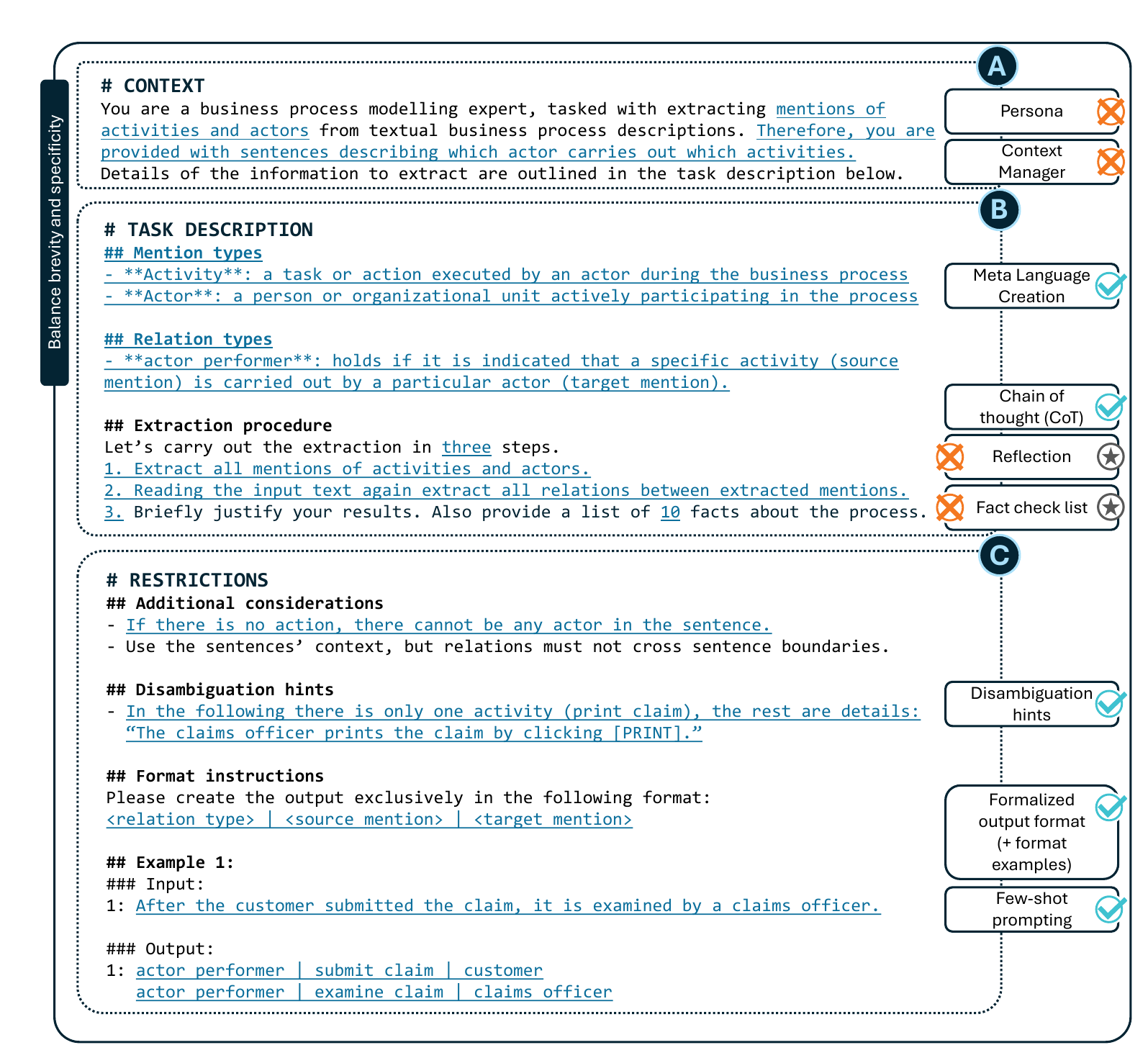}
    \caption{Modular prompt structure (underlined = task-specific content, boxes = design pattern, \ablationresultmarker{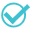} = useful, \ablationresultmarker{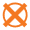} = non-useful, \ablationresultmarker{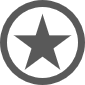} = use in prompt engineering).\vspace{-.5cm}}
    \label{fig:prompt-structure-overview}
\end{figure}

LLMs take freely formulated text as input, which is called the prompt. 
To this end we base our prompts on an ablation study (Section~\ref{sec:ablation}), which is used to identify beneficial and detrimental prompt components. To do this, we first need a modular prompt design so that we can specifically remove individual components in the study to examine their benefits and ultimately to only keep the advantageous components.
Adhering to the best practices outlined in \cite{tornberg2024bestpractices}, our initial prompt design is structured into three modules (see Figure~\ref{fig:prompt-structure-overview}): \textit{(A)} a context description framing the process information extraction task on a high level, \textit{(B)} a detailed task description, and \textit{(C)} constraints that further restrict the context and the output format, and contains disambiguation hints. To design potentially relevant components for all three modules we rely on general design patterns    \cite{min2023recent,white2023promptpatterncatalog,tornberg2024bestpractices,cui2021nerwithbart}.
Therefore, in the next subsection, the three modules are specified and discussed in terms of how they address the LLM-specific challenges outlined in Section~\ref{sec:challenges-llm}. 

\subsubsection{Context and Task Description.}\label{sec:prompting-context-and-task}

In module \textit{(A)} we use the \textit{persona} design pattern~\cite{white2023promptpatterncatalog} to control the language style of generation results. We assign it the role of a process modeling expert. This is followed by the \textit{context manager}\label{bp:context-manager} design pattern~\cite{white2023promptpatterncatalog}, which includes a general description of the information extraction task (i.e., objectives and a description of the input specifics). This limits the information basis the LLM may use, mitigating the risk of hallucination.  

Module \textit{(B)} is mainly concerned with defining the specifics of the process information extraction task. Its backbone is the \textit{creation of a meta language}\label{bp:meta-language}\cite{white2023promptpatterncatalog}, which defines the types of elements to extract from the input text. Figure~\ref{fig:prompt-structure-overview} provides an example that defines activities and actors as mention types and a relation called \textit{actor performer} that associates actions with their performers, following our running example from Figure~\ref{fig:running-example}. Another widespread best practice is known as \textit{chain of thought (CoT)}\label{bp:cot}~\cite{wei2022chain,tornberg2024bestpractices}, whereby the actual task is broken down into individual steps. Thus, our prompt divides the relation extraction task into two steps that separate the extraction of mentions from the prediction of their relations and a third step, which combines two more best practices, i.e., generating a \textit{list of facts} about the process\label{bp:facts} and \textit{reflection}\label{bp:reflection} about the results~\cite{white2023promptpatterncatalog}. These cause the LLM to elaborate both on the input and on its own output, which allows experts to validate the extracted information and has also been shown to have a positive impact on extraction performance~\cite{wei2022chain,tornberg2024bestpractices}.

\subsubsection{Restrictions.}\label{sec:prompting-restrictions}

The last prompt module \textit{(C)} defines expectations towards the LLM's output.
\textit{Additional considerations} include rules for the extraction task, such as that an actor can only be described as such if the action it performs is also named (compare Figure~\ref{fig:prompt-structure-overview}).
\textit{Disambiguation hints}\label{bp:disambiguation} are particularly useful for information types that are hard to distinguish from other information types or irrelevant information. In Figure~\ref{fig:prompt-structure-overview} it is intended to guide the LLM what makes up an \textit{Activity}, if the input gives additional, irrelevant specifics.  
Prompts further include a schematic definition of a \textit{formal output format}\label{bp:format}~\cite{white2023promptpatterncatalog}, which for the exemplary prompt is a tuple of a relation type, a source mention and a target mention, each separated by a pipe symbol. The definition is complemented by an (out-of-domain) example.
Finally, \textit{few-shot prompting}\label{bp:few-shots}~\cite{min2023recent,tornberg2024bestpractices} dynamically adds examples for input and corresponding output. For the current paper few-shot samples are pairs of raw textual process descriptions and a task-dependent set of process-relevant information (e.g., actor mentions). This design pattern and best practice is known to alleviate the issue of \textit{data unawareness} of LLMs~\cite{kaddour2023challenges,lewis2020retrieval}.

\subsection{Process Model Generation - a Proof of Concept}\label{sec:model-generation}

Based on the format instructions contained in the prompt design 
(Section~\ref{sec:prompting}), information can be extracted automatically 
through a parser. This information can then be automatically converted into
a process model in BPMN language, which was chosen because it is a widely used industry standard. 

\begin{figure}[bt]
\centering
\begin{subfigure}[t]{\textwidth}
    \centering
    \includegraphics[width=\linewidth]{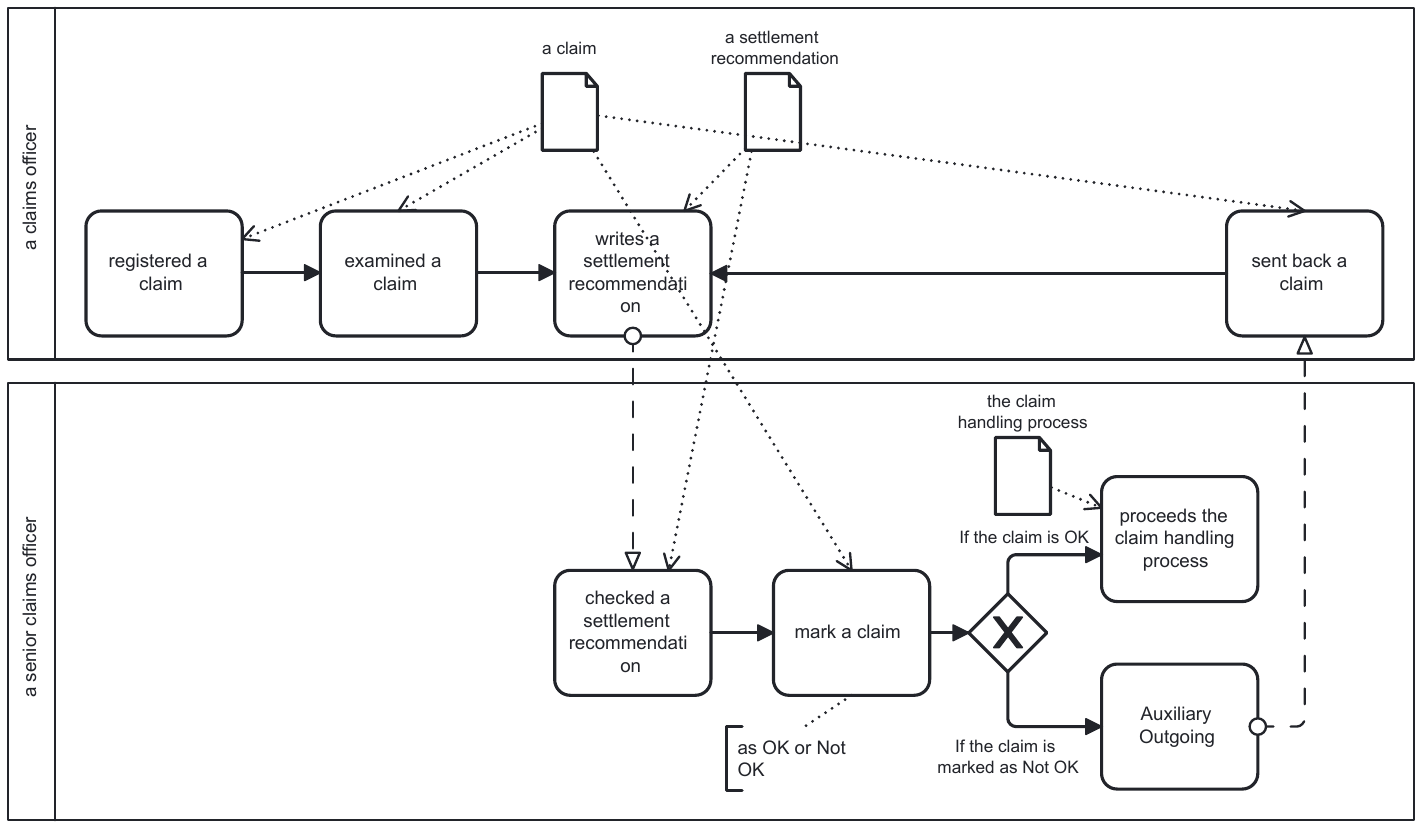}
\end{subfigure}%
\caption{Automatically generated model for information in document 
\emph{doc-3.3} (PET). Layout has been adjusted slightly for
readability and compactness.
\vspace{-.5cm}}
\label{fig:auto-generated-model}
\end{figure}

Our conversion algorithm works in three major stages. First, in the 
\emph{\textbf{Consolidation}} phase, we assign the conditions to the mentions
of their respective decisions. Next, all mentions of gateways are merged, if
they refer to the same decision point. Finally, for all activities that are 
not assigned an executing actor, we find the closest actor mention in the 
text left of it.
In the second stage, the \emph{\textbf{Vertex}} stage, we create process elements for
all mentions, e.g., \emph{Tasks}, \emph{Data Objects}, \emph{Swimlanes}, etc.
The final \emph{\textbf{Linking}}
stage connects related elements, e.g., successive tasks with 
\emph{Sequence Flows}, if they are located in the same Swimlane, or 
\emph{Message Flows} otherwise. We also create \emph{Data Associations} between 
Data Objects and Tasks, adding the label of the Data Object to 
the label of the Task, for labels like ``\emph{register the 
claim}''. 

Figure \ref{fig:auto-generated-model} shows the resulting BPMN model after
applying this algorithm on the information contained in document 
\emph{doc-3.3} of the (\hyperref[data:pet]{PET}) dataset we later use in our experiments. Note, that the layouting is adjusted
manually, as well as the wrong assignment of Task \emph{sent back a claim} to 
the Swimlane of Actor \emph{A claims officer}. These issues illustrate why 
synthesizing a model from extracted process information is still challenging, which warrants further research beyond the scope of this paper.

\section{Experiment Setup}\label{sec:experiment_setup}

In this section we define the experiments we run to evaluate the
usefulness of LLMs for process information extraction
\footnote{Source code: \url{https://anonymous.4open.science/r/llm-process-generation-2140} and \url{https://anonymous.4open.science/r/pet-to-bpmn-poc-B465}.}. 
It covers 
an overview of the datasets we use, including the respective baselines, 
and a definition of metrics we apply.


We use three well-known datasets for evaluating our prompts. One of these 
(\hyperref[data:pet]{PET}) represents the current state of the art, both in terms of size, as well 
as the process information techniques developed for it. The other datasets
feature different characteristics, making them relevant for validation 
experiments. This lets us gain insights into the robustness of an LLM as a 
process information extractor, as well as how it behaves when applied to 
other process modeling languages. We call the best approaches for
extracting the information from these datasets \textit{baselines}, and 
use them in Section~\ref{sec:results} as comparisons with various LLMs.

\noindent \textbf{PET}~\cite{bellan2023pet}\label{data:pet}: This is the largest dataset 
currently available. It contains 45 documents with annotations for information 
especially useful for creating process models in BPMN. These include 7 types of 
mentions such as activities, actors, data objects, but 
also 6 relation types. These cover the behavioural process perspective (\textit{Flow}), 
data perspective (\textit{uses}, and organizational 
perspective (\textit{actor, performer}). Additionally, this dataset features 
relations that span multiple sentences. It therefore tests the ability of 
approaches to reason across wider
spans of text. We use an extended version of this dataset, which includes data for 
the \hyperref[task:er]{ER} task, as presented in~\cite{neuberger2023beyond}.
The currently best approach for extracting information is using conditional random fields for \hyperref[task:md]{MD}, 
a pre-trained neural co-reference resolver for \hyperref[task:er]{ER}, 
and a decision tree ensemble for \hyperref[task:re]{RE}~\cite{neuberger2023beyond}.
We use scores as reported in~\cite{neuberger2023beyond}, which have corresponding
publicly available code and can be reproduced with it.

\noindent \textbf{DECON}~\cite{van2019extracting}\label{data:declare}: This collection 
of 17 textual process descriptions is annotated with a set of 5 \underline{De}clare~\cite{pesic2007declare} 
\underline{Con}straint types between business process relevant
activities. Additionally, constraints may be negated, as 
well as unary, i.e., constraining a single action. Annotations are 
given on a sentence level, and only
sentences that describe at least one constraint are contained in the 
dataset. The expected (ground-truth) activities are already transformed
into Declare-conform phrases, i.e., the activity description ``\textit{%
The claim is registered}'' should be extracted as ``\textit{register 
claim}''. It does not contain an approach for \hyperref[task:md]{MD} 
in isolation, and only contains mentions of type action. The authors
of~\cite{van2019extracting} propose a rule-based approach combining
multiple NLP techniques, e.g., \textit{typed dependency relations}.
    
\noindent \textbf{ATDP}~\cite{quishpi2020extracting}\label{data:atdp}: This dataset uses 
18 textual descriptions, that largely overlap with the ones from
\cite{van2019extracting}, but also contains sentences, that 
describe no constraint. As such, this dataset tests approaches for 
their ability to judge whether or not sentences contain relevant information, 
before extracting constraints. Furthermore, the set of constraints was 
expanded to eight types. Additionally, this dataset also provides
annotations of actions, conditions, entities, and events, which we used 
in an \hyperref[task:md]{MD} setting. Quishpi et al. proposed
a rule-based ensemble of patterns for \hyperref[task:md]{MD} and 
\hyperref[task:ce]{CE} on typed dependency structures~\cite{quishpi2020extracting}.



We use the well established metrics \textit{Precision} $P$ and \textit{Recall} $R$
for our experiments. $P$ is a measure of how well an extraction approach is able
to avoid false positives, i.e., assigning the wrong type to mentions and relations,
or extracting them, where they are not expected. $R$ on the other hand measures how
much of the expected information (true positives) is found. The two
metrics are typically aggregated via their harmonic mean 
$F_1 = 2\frac{P \cdot R}{P + R}$.
Following~\cite{quishpi2020extracting}, we use $P=\frac{\#correct}{\#pred}$ and 
$R=\frac{\#correct}{\#gold}$, with $\#correct$ as the number of correct predictions,
$\#pred$ the number of total predictions, and $\#gold$ as the number of expected mentions or relations. For a fair assessment, we count predictions as correct in 
exactly the way described by the work we compare the LLM to.

\section{Results}\label{sec:results}

Table~\ref{tab:detailed_results} shows the results we observed when running the experiments 
as described in Section~\ref{sec:experiment_setup} with an optimized prompt, that follows 
the recommendations we found in our study of best practices (Section~\ref{sec:ablation}). 
All results use GPT-4o, the latest version of OpenAI's GPT with $temperature=0$. $top\_p$ is unchanged, as per OpenAI's recommendation, when using temperature based sampling.
\footnote{see \href{https://github.com/openai/openai-python/blob/main/src/openai/resources/chat/completions.py\#L196}{OpenAI's source code}, accessed June 3, 2024}

For the reference dataset \hyperref[data:pet]{PET}, our experiments show that GPT-4o is 
capable of an absolute $F_1$ score improvement of 5\% for \hyperref[task:md]{MD}, 22\% 
for \hyperref[task:er]{ER}, and 17\% for \hyperref[task:re]{RE}. Remarkably, for \hyperref[task:re]{RE}, 
GPT-4o 
is able to match and outperform the machine learnt baseline, which was trained on 36 manually 
annotated documents~\cite{neuberger2023beyond}, without any labeled data (zero-shot). For 
\hyperref[task:md]{MD} it reaches similar scores, even when not given any examples, compared to the machine 
learnt baseline, which was trained using 36 manually annotated documents~\cite{bellan2023pet}.
For real-world application this means that LLMs can be used in business process information extraction
scenarios, even if the organization has not a single manually annotated training example.
This is an exciting find, as it promises significant speed-up of model creating tasks of
practitioners across business domains.

When evaluating on the validation datasets (cf. Section~\ref{sec:experiment_setup}), we found 
that GPT-4o is able to match and out-perform the rule-based systems in all cases, most notably improving 
$F_1$ scores for \hyperref[task:re]{RE} on dataset \hyperref[data:declare]{DECON} by 
an absolute 8\%. Our result for \hyperref[task:md]{MD} on dataset \hyperref[data:declare]{DECON} has no corresponding
baseline, as the authors of~\cite{van2019extracting} did not report values for \hyperref[task:md]{MD} in isolation.
Errors and ambiguities are common in dataset \hyperref[data:atdp]{ATDP} hindering machine learning methods in learning 
valid extraction rules. This also adversely affects the extraction accuracy of GPT-4, when extracting the same types 
of constraints in the \hyperref[data:atdp]{ATDP} dataset compared to the \hyperref[data:declare]{DECON} dataset. 
We discuss this further in Section~\ref{sec:lessons-learned}.
Since the importance of \hyperref[task:er]{ER} only recently gained 
attention~\cite{quishpi2020extracting,neuberger2023beyond}, the reference dataset 
\hyperref[data:pet]{PET} is currently the only dataset providing data for evaluation of this
task.

\begin{table}[tb]
    \centering
    \begin{tabularx}{\textwidth}{r|r|CCC|CCC|CCC|}
        & Dataset& \colheader{\hyperref[data:declare]{\textbf{DECON}}} & \colheader{\hyperref[data:atdp]{\textbf{ATDP}}} & \colheader{\hyperref[data:pet]{\textbf{PET}}} \\
        & Metric& $P$ & $R$ & $F_1$ & $P$ & $R$ & $F_1$ & $P$ & $R$ & $F_1$ \\
        \hlinex{1.25pt}
        \rowheader{4}{Mention\\Detection}   & Baseline          & \multicolumn{3}{c|}{\cellcolor[gray]{.95}\textit{no baseline}} & 0.62 & \textbf{0.82} & 0.71          & \textbf{0.73} & 0.64 & 0.69 \\
        \cline{2-11}
                                            & Zero-shot         & 0.72 & 0.75 & 0.73                            & 0.58 & 0.77 & 0.66                            & 0.65 & 0.71 & 0.68 \\
                                            
                                            & 1-shot            & 0.87 & \textbf{0.80} & 0.83                   & 0.63 & 0.77 & 0.69                            & 0.72 & 0.75 & 0.73 \\
                                            
                                            & 3-shot            & \textbf{0.88} & 0.79 & \textbf{0.83}          & \textbf{0.68} & 0.79 & \textbf{0.73}          & 0.72 & \textbf{0.77} & \textbf{0.74} \\
                                            
        \hlinex{1.25pt}
        \rowheader{4}{Entity\\Resolution}   & Baseline          & \multicolumn{3}{c|}{\cellcolor[gray]{.95}}      & \multicolumn{3}{c|}{\cellcolor[gray]{.95}}      & 0.55 & \textbf{0.51} & 0.52 \\
        \cline{2-2}\cline{9-11}
                                            & Zero-shot         & \multicolumn{3}{c|}{\cellcolor[gray]{.95}} & \multicolumn{3}{c|}{\cellcolor[gray]{.95}}      & 0.67 & 0.55 & 0.60 \\
                                            
                                            & 1-shot            & \multicolumn{3}{c|}{\cellcolor[gray]{.95}} & \multicolumn{3}{c|}{\cellcolor[gray]{.95}}      & 0.76 & \textbf{0.70} & 0.73 \\
                                            
                                            & 3-shot            & \multicolumn{3}{c|}{\multirow{-4}{*}{\cellcolor[gray]{.95}\textit{no data}}} & \multicolumn{3}{c|}{\multirow{-4}{*}{\cellcolor[gray]{.95}\textit{no data}}}      & \textbf{0.79} & \textbf{0.70} & \textbf{0.74} \\
                                            
        \hlinex{1.25pt}
        \rowheader{4}{Relation\\Extraction} & Baseline         & 0.77 & 0.72  & 0.74    & \textbf{0.58}  & 0.64  & 0.61    & 0.79 & 0.66 & 0.72 \\
        \cline{3-8}
        \cline{2-11}
                                            & Zero-shot        & 0.66 & 0.75  & 0.70                                    & 0.49  & 0.66 & 0.57                                           & 0.88 & 0.85 & 0.86 \\
                                            
                                            & 1-shot           & 0.76 & 0.82  & 0.79                                    & \textbf{0.58}  & \textbf{0.73}  & \textbf{0.64}               & 0.90 & 0.89 & 0.89 \\
                                            
                                            & 3-shot           & \textbf{0.79} & \textbf{0.85}  & \textbf{0.82}         & \textbf{0.58}  & 0.72           & \textbf{0.64}               & \textbf{0.90} & \textbf{0.89} & \textbf{0.89} \\
    \end{tabularx}
    \caption{Results for each dataset and the different extraction stages, 
    compared to baseline results using GPT-4o.\vspace{-.5cm}}
    \label{tab:detailed_results}
\end{table}

\subsection{Model Comparison}\label{sec:model-comparision}

We originally developed our prompts for GPT-4 version \emph{GPT-4-0125-preview},
to assess how well our prompting strategy generalizes to other models
we prompted a total of eight models for the MD and RE tasks on PET.
We selected models following AlpacaEval
\footnote{see \url{https://tatsu-lab.github.io/alpaca_eval/}, last accessed May 30, 2024.}, 
which is designed for testing the instruction following capabilities of 
LMMs\cite{dubois2024alpacafarm}. At the time of writing model 
\emph{YI Large Preview} was not publicly accessible and could not be
considered in our comparison, even though it ranked third on AlpacaEval.

Results for the comparison can be found in Table~\ref{tab:model-comparison}.
We set the temperature for all models to $0$.
All GPT models perform on similar levels, with the exception of GPT3.5,
which is significantly smaller compared to GPT-4 models. For the zero-shot
RE task GPT3.5 even failed to produce responses for most documents, leading
to very low recall. Claude3 Opus seems to be as capable as GPT-4, its smaller
variant Sonnet performs significantly worse on zero-shot tasks, but is able
to produce comparable results given three examples. Llama 3 70B Instruct is
an open-weight model and could be run locally, i.e., it is useful for using
our prompting strategy in scenarios where sending data to an API is not 
possible. Llama 3 70B Instruct seems to be nearly as capable as the 
closed-weights Claude 3 Sonnet and is therefore viable in a few-shot setting.

\begin{table}[tb]
    \centering
    \begin{tabularx}{\textwidth}{r|CCC|CCC|CCC|CCC|}
        Task & \colheader{\hyperref[data:pet]{\textbf{PET}} \hyperref[task:md]{MD}\\(Zero-shot)} & \colheader{\hyperref[data:pet]{\textbf{PET}} \hyperref[task:md]{MD}\\(3-shot)} & \colheader{\hyperref[data:pet]{\textbf{PET}} \hyperref[task:re]{RE}\\(Zero-shot)} & \colheader{\hyperref[data:pet]{\textbf{PET}} \hyperref[task:re]{RE}\\(3-shot)} \\
        \hlinex{1pt}
        Model & $P$ & $R$ & $F_1$ & $P$ & $R$ & $F_1$ & $P$ & $R$ & $F_1$ & $P$ & $R$ & $F_1$ \\
        \hlinex{1.25pt}
        GPT-4o & 0.58 & 0.69 & 0.63 & 0.68 & 0.77 & 0.72 & \textbf{0.88} & \textbf{0.85} & \textbf{0.86} & 0.90 & 0.89 & 0.89 \\
        GPT-4-2024-04-09 & 0.63 & 0.67 & 0.65 & \textbf{0.73} & 0.76 & \textbf{0.74} & 0.87 & 0.79 & 0.83 & 0.89 & 0.88 & 0.88 \\
        GPT-4-0125-preview & \textbf{0.65} & 0.71 & \textbf{0.68} & 0.72 & 0.77 & \textbf{0.74} & 0.87 & \textbf{0.85} & \textbf{0.86} & 0.89 & 0.87 & 0.88 \\
        GPT-3.5-0125 & 0.35 & 0.50 & 0.42 & 0.51 & 0.70 & 0.59 & 0.51 & 0.06 & 0.11 & 0.74 & 0.64 & 0.69 \\
        \hlinex{0.75pt}
        Claude 3 Opus & 0.55 & \textbf{0.72} & 0.63 & 0.66 & \textbf{0.80} & 0.73 & 0.86 & \textbf{0.85} & \textbf{0.86} & \textbf{0.92} & \textbf{0.91} & \textbf{0.91} \\
        Claude 3 Sonnet & 0.46 & 0.65 & 0.54 & 0.63 & 0.78 &0.70 & 0.78 & 0.67 & 0.72 & 0.91 & 0.87 & 0.89 \\
        \hlinex{0.75pt}
        Llama 3 70B Instruct & 0.56 & 0.64 & 0.59 & 0.62 & 0.71 & 0.67 & 0.76 & 0.66 & 0.70 & 0.88 & 0.81 & 0.84 \\
        Qwen1.5 72B Chat & 0.32 & 0.33 & 0.33 & 0.53 & 0.65 & 0.59 & 0.61 & 0.65 & 0.63 & 0.74 & 0.77 & 0.75
    \end{tabularx}
    \caption{Comparison of our prompts across different models. Best results per task and metric are set \textbf{bold}. \label{tab:model-comparison}}
\end{table}

\subsection{Ablation Study}\label{sec:ablation}

We conduct an ablation study to assess the usefulness of the best practices presented
in Section~\ref{sec:prompting} and to measure the impact of the prompt's main components. 
This study is run on the reference dataset (\hyperref[data:pet]{PET}), as it is
the largest one and used by recent publications
\cite{bellan2023pet,bellan2022extracting,neuberger2023beyond}. 
To obtain a baseline for the tasks of \hyperref[task:md]{MD} and \hyperref[task:re]{RE},
we use a prompt that implements the best practices as shown in Figure
\ref{fig:prompt-structure-overview} and run it on the \emph{GPT-4-0125-preview} model. 
We then purposefully remove specific components from this prompt, namely the \textit{format examples}, the \textit{context manager}, the \textit{persona}, the definition 
of mention and relation types (\textit{meta language}), the instruction to think in several 
steps (\textit{chain of thought}), any \textit{disambiguation} hints, and the instruction 
to generate explanations (\textit{reflection}) and a \textit{fact check list} about the process. 
Additionally, we also use a prompt with very short descriptions of relations and types 
(\textit{balancing brevity and specificity}). We run each prompt in the zero-shot setting and 
record the observed $F_1$ score, as well as the parsing errors that occurred. 

Table \ref{tab:ablation-study-details} provides detailed results.
Removing examples 
has a 
significant negative effect ($-0.22$ for \hyperref[task:md]{MD} and $-0.07$ for \hyperref[task:re]{RE}), mainly rooted 
in the number of parsing errors that are made (919 for \hyperref[task:md]{MD}), 
as well as directionality of relations for the \hyperref[task:re]{RE} task (confusing 
source and target mentions). Removing the context manager 
and persona only has minor effects ($\pm0.01$ per task), suggesting lower relevance
for process information extraction  compared to other NLP settings.

\begin{table}[bt]
    \sisetup{table-align-text-post=false, table-space-text-post={\,\%}}

    \centering
    \begin{tabularx}{\textwidth}{LS[table-format=1.2,print-implicit-plus]S[table-format=1.2]S[table-format=3]S[table-format=1.2,print-implicit-plus]S[table-format=1.2]S[table-format=3]c}
        & \multicolumn{3}{c}{Mention Detection (\hyperref[task:md]{MD})} & \multicolumn{3}{c}{Relation Extraction (\hyperref[task:re]{RE})} \\
        \cmidrule(r){2-4} \cmidrule(l){5-7}
         {\thead[l]{Experiment}}        & {\thead{Relative\\$F_1$}}   & {\thead{Absolute\\$F_1$}}   & {\thead{Parsing\\Errors}} & {\thead{Relative\\$F_1$}}   & {\thead{Absolute\\$F_1$}}   & {\thead{Parsing\\Errors}} & {\thead{Useful}}   \\
         
         \toprule
         Baseline                   & {--}                      & 0.59                      & 0     & {--}  & 0.77 & 0 &                    \\   
         
         No Format Examples         & -0.22                     & 0.37                      & 919   & -0.07 & 0.70 & 1 & \checkmark \\
                  
         No Context Manager         & +0.01                     & 0.60                      & 0     & -0.01 & 0.76 & 0 & \\
         
         No Persona                 & +0.01                     & 0.59                      & 1     & +0.01 & 0.78 & 0 & \\

         No Meta Language           & -0.09                     & 0.49                      & 2     & -0.05 & 0.72 & 0 & \checkmark \\

         No Chain of Thought        & -0.01                     & 0.57                      & 1     & -0.02 & 0.75 & 0 & \checkmark \\

         No Disambiguation          & -0.03                     & 0.55                      & 0     & -0.01 & 0.76 & 1 & \checkmark \\

         No Reflection              & +0.04                     & 0.63                      & 0     & +0.02 & 0.79 & 0 & $\bigstar$ \\

         No Fact Check List         & +0.03                     & 0.62                      & 1     & -0.02 & 0.75 & 0 & $\bigstar$ \\

         Very Short Prompt          & -0.04                     & 0.54                      & 1     & -0.03 & 0.74 & 0 & \checkmark \\
         
         \bottomrule
    \end{tabularx}
    \caption{Changes in $F_1$ score of GPT-4, without specific prompt components given in
    Figure~\ref{fig:prompt-structure-overview}. Column \textit{relative $F_1$} shows difference to the 
    baseline prompt, \textit{Useful} shows a \checkmark, if we recommend this component in prompts for process 
    information extraction and $\bigstar$ for prompt engineering and data curating only.\vspace{-.6cm}}
    \label{tab:ablation-study-details}
\end{table}

In addition to removing prompt components, we also tested using GPT in an older, 
less capable, but much cheaper version, GPT-3.5. Running the baseline prompt, 
results in a significant drop in extraction quality, with $F_1=0.27$ for \hyperref[task:md]{MD} and
$F_1=0.56$ for \hyperref[task:re]{RE}. Splitting the baseline prompt into multiple prompts, each focusing
on only one mention type, lets us prompt GPT-4 repeatedly for the same document. 
These highly specialized prompts are called ``agents'', which pass information 
between each other. For example, we instruct the first agent to extract \textit{Actions}, 
which are passed to other agents extracting \textit{Actors} and \textit{Business
Objects} respectively. This lets us exploit the inherent dependency between these
elements. If no Action is detected in a sentence, then there is likely no relevant
Actor or Business Object, even if there are nouns that would qualify from a 
linguistic standpoint. This way of prompting leads to an absolute improvement of
$+0.08$ in $F_1$ score for both the \hyperref[task:md]{MD} and \hyperref[task:re]{RE} 
tasks.

\subsection{Lessons Learned}\label{sec:lessons-learned}

Using LLMs for extracting process relevant information brings with it a
category of challenges, which we already discussed in Section~\ref{sec:challenges-llm}.
Solving these is paramount for successful application of LLMs. In this section we 
discuss how we approached these challenges and what lessons we learned.

\noindent \textbf{\hyperref[challenge:output-control]{(\textit{C1})} Limited output control}. The expected output format, as 
well as the form of extracted information, can mainly be influenced by the
prompt components \textit{Meta Language} and \textit{Format Examples}.
Adding these results in significantly improved $F_1$ scores, 
($+0.22$ and $+0.05$ respectively for \hyperref[task:md]{MD} on \hyperref[data:pet]{PET}). 
These improvements are explained by less parsing 
errors (919 less for \hyperref[task:md]{MD} on \hyperref[data:pet]{PET}), and better recall and precision in detecting
mentions. LLMs also run the risk of being ``stochastic parrots'', simply 
synthesizing linguistically correct phrases, based on their training data
\cite{bender2021dangers}. In our experiments we observed changes in
$F_1$ of maximally $-0.02$ for rephrased prompts. This indicates robustness of 
our prompts and suitability of LLMs as an tool for business process 
information extraction.

\begin{figure}[tb]
\centering
    \centering
    \includegraphics[width=\linewidth]{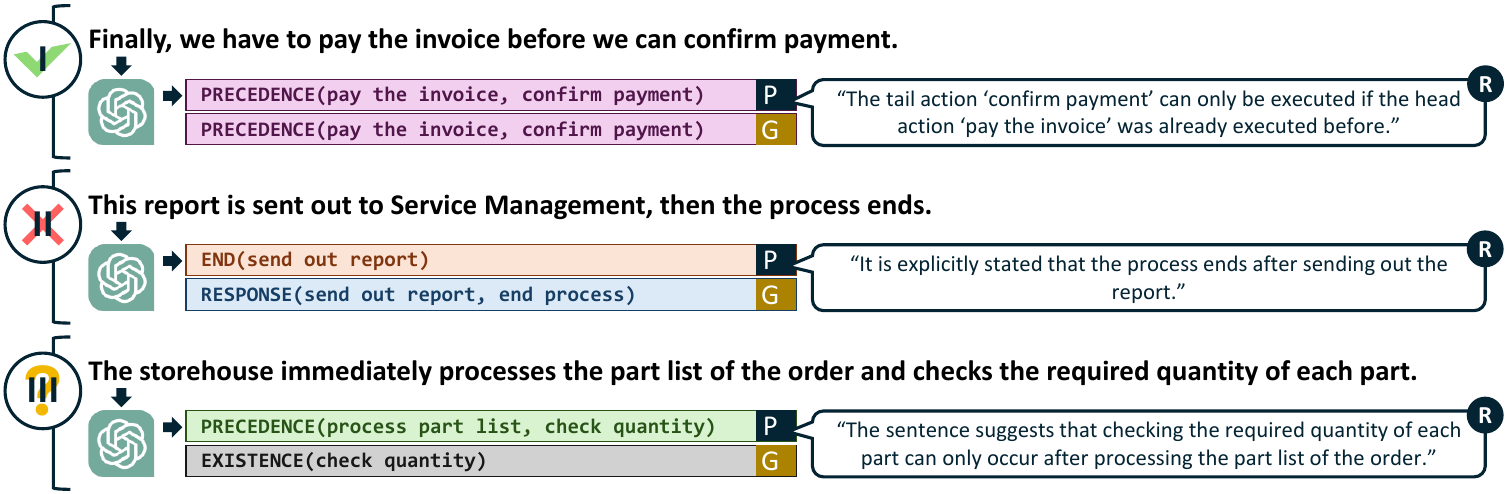}
    \caption{\underline{R}eflection example with \underline{P}redicted and \underline{G}old standard constraints: (I) perfect match, (II) gold standard error, (III) ambiguous case}
    \label{fig:llm-explanations}
    \vspace{-.5cm}
\end{figure}

\noindent \textbf{\hyperref[challenge:black-box]{(\textit{C2})} Black-box}. A valuable advantage of utilizing LLMs is their ability to reflect, thereby providing justification for their generated results. 
%
%
Figure~\ref{fig:llm-explanations} shows three examples of justifications for extraction results in the \hyperref[data:atdp]{ATDP} dataset (see Section~\ref{sec:task-descriptions}). 
Case I shows the ideal outcome where prediction and expected constraint are identical. 
Note, that the justification even refers to the meta-language provided in the prompt (compare Section~\ref{sec:prompting}). 
In case II, the prediction and the gold standard constraint do not match, because of an error in the gold standard data, following \cite{van2019extracting}, which defines completing a process as a \emph{meta action} that can not be part of any constraint. The dataset creators are alerted of this issue by the LLM, since it plausibly justifies why \emph{send out report} marks the end of a process instance.
Finally, in case III the extraction result is controversial, since the sentence is 
ambiguous. If we consider the term \emph{immediately} to encompass both actions, they are constrained by an \emph{existence} constraint. Alternatively, viewing \emph{check quantity} as a subtask of \emph{process part list} suggests only one action needs modeling. 
Using such reflective explanations make LLMs useful for ``human-in-the-loop'' 
systems, which are already applied in fields like process mining
\cite{ter2023process}.

\noindent \textbf{\hyperref[challenge:input-presentation]{(\textit{C3})} Input presentation dependencies}. Adding more text to prompts sometimes has an adverse effect, reducing extraction 
quality (Section~\ref{sec:challenges-llm}). This makes optimizing prompts difficult, since it is not clear, if adding 
additional disambiguation hints or longer definitions would improve the result. 
Using partial extraction prompts helps 
with this issue, as the sections regarding \textit{Meta Language Creation} 
can be focused on a few types. Depending on the task, there may even be inter-dependency 
between information, that can be efficiently exploited in this way.

\noindent \textbf{\hyperref[challenge:data-unawareness]{(\textit{C4})} Data unawareness}. This issue arises, when LLMs are used 
in a zero-shot setting. There, the components \textit{Meta Language},
and \textit{Format Examples} are the only ways to ``teach'' the model how 
to perform the task. Applying the pattern of \textit{few-shot prompting}, 
i.e., using labeled data in a few-shot setting was beneficial. This makes the use 
of an LLM more akin 
to training a machine learning model, but with significantly lower data requirements.
In our experiments, three examples were sufficient to achieve better
extraction results than those of machine learning models trained with more than ten times
of the data.

\noindent \textbf{\hyperref[challenge:costly-experiments]{(\textit{C5})} Costly experiments}. 
This is a major drawback of LLM based process information extraction. The most 
capable LLMs are hosted as cloud-based solutions
and are priced per token. We found that limiting the number
of examples to 1 resulted in the best cost-value ratio. Additionally,
our experiments showed that leaving out the prompt components \textit{Context 
Manager}, \textit{Persona}, and \textit{Disambiguation} is a valid way to limit
the number of tokens sent per request, albeit with potential minor decreases in 
extraction accuracy. Prompting LLMs without the request
for a \textit{Fact List}, nor \textit{Explanations} for extracted
information greatly reduces the amount of tokens as well, especially useful
after prompt engineering or data curating (during ``\textit{inference}'').
Alternatively one can switch to cheaper models, i.e., \textit{LLama 3 72B},
if the drop in performance is acceptable.

\section{Conclusion}\label{sec:conclusion}

\textbf{Summary}. This paper presents an extensive study on the usefulness of 
LLMs for the extraction of  process
 information from natural language text. We collected linguistic challenges and discuss how LLMs are uniquely fit for
solving them. We also discussed  
challenges that arise through the use of LLMs and show how
other communities propose to deal with these (or similar) concerns through prompt engineering. We present experimental results on three process information extraction datasets, which at least match the current state of the art on these datasets and in 
most cases improve it by as much as 8\%
in the $F_1$ metric. This shows the suitability of LLMs as a method
for extracting business process relevant information from natural 
language process descriptions. To flesh out this notion, we analyze
how well our prompting strategy can be applied to different LLMs without
changing them, showing their universal nature. 
We make all
our code, prompts, and LLM answers available,
to support further research.

\noindent\textbf{Limitations}. A limitation of our work is that the list of prompt components we present may not be 
exhaustive, and they may have interactions that our ablation study 
does not capture.
Additionally, some models suffer from hallucinations, especially Qwen1.5 and 
Llama 3, which hallucinate non-existent entity and relation types -- 20 and 37 
instances in the worst cases respectively. However, the severity of this problem 
diminishes in the few-shot setting (0 and 4 instances respectively in the 
worst case). We plan on analyzing how our prompts could be enhanced to 
improve instruction following for these models.
Lastly, the current pricing models prohibit large-scale application of the most
capable model to process information extraction. Alternatives (e.g., Llama) avoid this issue, but require more labeled examples to reach comparable
levels of performance.

\noindent\textbf{Future Work}. In future work we aim to use LLMs as tools
to support labeling of new data. Current datasets are limited in origin, i.e.,
they usually describe processes from municipalities or small service 
providers. We plan to analyze the ability of LLMs to generalize beyond the domains with available labeled data and highlight the promising flexibility observed in our current experiments. Finally, using GPT-4o's image processing and generation capabilities could be a promising line of research for direct text to model transformation.

\bibliographystyle{splncs04}
\bibliography{bibliography}

\end{document}